# TWO-TIMESCALE LEARNING USING IDIOTYPIC BEHAVIOUR MEDIATION FOR A NAVIGATING MOBILE ROBOT

**Amanda M. Whitbrook, Uwe Aickelin, Jonathan M. Garibaldi**

Intelligent Modelling and Analysis Research Group,

School of Computer Science, University of Nottingham, U.K.

{amw, uxa, jmg}@cs.nott.ac.uk

## ABSTRACT

A combined Short-Term Learning (STL) and Long-Term Learning (LTL) approach to solving mobile-robot navigation problems is presented and tested in both the real and virtual domains. The LTL phase consists of rapid simulations that use a Genetic Algorithm to derive diverse sets of behaviours, encoded as variable sets of attributes, and the STL phase is an idiotypic Artificial Immune System. Results from the LTL phase show that sets of behaviours develop very rapidly, and significantly greater diversity is obtained when multiple autonomous populations are used, rather than a single one. The architecture is assessed under various scenarios, including removal of the LTL phase and switching off the idiotypic mechanism in the STL phase. The comparisons provide substantial evidence that the best option is the inclusion of both the LTL phase and the idiotypic system. In addition, this paper shows that structurally different environments can be used for the two phases without compromising transferability.

# 1     INTRODUCTION

An important decision when designing effective controllers for mobile robots is how much *a priori* knowledge should be imparted to them. Should they attempt to learn all behaviours during the task, or should they begin with a set of pre-engineered actions? Both of these alternatives have considerable drawbacks; starting with no prior knowledge increases task time substantially because the robot has to undergo a learning period during which it is also at risk of damage. However, if it is solely reliant on designer-prescribed behaviours, it has no capacity for learning and adaptation.

The architecture described in this paper takes inspiration from the vertebrate immune system in order to attempt to overcome these problems. The immune system learns to recognize antigens over the lifetime of the individual, which constitutes Short-Term Learning (STL), but it also knows how to build successful antibodies from gene libraries that have evolved over the lifetime of the species. This represents Long-Term Learning (LTL), defined as that which evolves and develops as a species interacts with its environment and reproduces itself. Here, this "two timescale" approach is mimicked by using an Artificial Immune System (AIS) to represent the STL phase, and a Genetic Algorithm (GA) to represent the LTL phase. The GA rapidly evolves sets of behaviours in simulation to seed the AIS, which removes the need for using pre-engineered behaviours, and prevents robots from having to begin a task with no knowledge. The GA and AIS are run consecutively, with the GA running first. Once the GA has converged, the evolved antibody information is stored in a database for AIS initialization, and the GA does not run again.

An idiotypic network that uses Reinforcement Learning (RL) to update antibody-antigen matching is selected for the AIS system, and Farmer's computational model (Farmer 1986) of Jerne's idiotypic-network theory (Jerne 1974) is adopted. This model uses the analogy of antibodies as robot behaviours and antigens as environmental stimuli, and, in theory, has great potential to create very flexible and dynamic robots that can adapt to their environment. However, most previous implementations of the method exhibit rather limited self-discovery and learning properties, since their designs use very small numbers of pre-engineered behaviours, and only the network connections between them are evolved. In the architecture described here, the actual behaviours themselves are evolved, which permits delivery of novel and diverse sets of antibodies for seeding the AIS.

This paper demonstrates the importance of seeding (i.e. including an LTL-phase) by comparing schemes that employ the GA with those that do not. It also investigates the benefits of idiotypic selection by comparing idiotypic systems with AIS schemes that rely on RL only. In addition, it examines whether antibody replacement is necessary in seeded AIS systems. Finally, as a result of all these investigations, an attempt is made to represent the antigen space (i.e. the environment) more fully. Here, comparisons with the previous results show that a more complex representation does not enhance performance.

## 2    BACKGROUND AND MOTIVATION

Throughout the lifetime of an individual, the adaptive immune system learns to recognize antigens by building up high concentrations of antibodies that have proved useful in the past, and by eliminating those deemed redundant. This is a form of STL. However, the antibody repertoire is not random at birth and the mechanism by which antibodies are replaced is not a random process. Antibodies are built from gene libraries that have evolved over the lifetime of the species. This suggests that the immune system depends on both STL and LTL in order to achieve its purpose.

AIS algorithms take inspiration from the natural immune system (de Castro and Timmis 2002), and a variety of different models have been applied to both hardware (Canham *et al.* 2003) and software (Neal and Timmis 2003) robotics. However, the most popular robotics software model has been the idiotypic network, based on Farmer's model of continuous antibody-concentration change. In this model the concentrations are not only dependent upon past matching to antigens, they also depend on the other antibodies present in the system, i.e. antibodies are continually suppressed and stimulated by each other as well as being stimulated by antigens. In theory this design permits great variability of robot behaviour since the antibody that best matches the invading antigen is not necessarily selected for execution; the complex dynamics of stimulation and suppression ensure that suitable alternative antibodies are tried when the need arises (see Whitbrook *et al.* 2007).

However, past work in this area has mostly focused on how the antibodies in the network should be connected and, for simplicity, has used a single set of pre-engineered behaviours for the antibodies, which limits the potential of the method. For example, Watanabe *et al.* (1998a and 1998b) use an idiotypic network to control a garbage-collecting robot. Their antibodies are composed of a precondition, a behaviour, and an idiotope part that defines antibody connection. However, the sets of possible behaviours and preconditions are fixed, and only the idiotope part is evolved. Michelan and Von Zuben (2002) and Vargas *et al.* (2003) also use GAs, but again only the idiotypic-network connections are derived. Krautmacher and Dilger (2004) apply the idiotypic method to robot navigation, but their emphasis is on the use of a variable set of antigens; they do not change or develop the initial set of handcrafted antibodies, as only the network links are evolved. Luh and Liu (2004) address target-finding using an idiotypic system, modelling their antibodies as steering directions. However, although many behaviours are technically possible since any angle can be selected, the method is limited because a behaviour is defined only as a steering angle. Hart *et al.* (2003) update their network links dynamically using RL and a skill hierarchy, so that more complex tasks are achieved by building on basic ones, but the initial behaviours are hand-designed at the start.

It is clear that the idiotypic AIS methodology holds great promise for providing a system that can adapt to change, but its potential has never been fully explored because of the limits imposed on the fundamental behaviour-set. This research aims to widen the scope of the idiotypic network by combining LTL with STL, as in the natural immune system. The LTL consists of a GA in which six basic antibody-types are encoded with a set of six variable attributes that can take many different values, meaning that the system can evolve complete sets of simple but very diverse antibodies. These can then be passed to the STL phase, as a form of seeding or intelligent initialization for the AIS. In addition, the seeding provides the potential to bestow much greater flexibility to the idiotypic system, as an evolved set of distinct behaviours is available for each known antigen, providing a degree of choice.

LTL in simulation coupled with an idiotypic AIS in the real world represents a novel combination for robot-control systems, and should provide definite advantages, not only for AIS initialization, but also for evolutionary robotics. In the past, much evolutionary work has been carried out serially on physical robots, which requires a long time for convergence and puts the robot and its environment at risk of damage. For example, Floreano and Mondada (1996) adopt this approach and report a convergence time of ten days. More recent evolutionary experiments with physical robots, for example Marocca and Floreano (2002,) Hornby *et al.* (2000), and Zykov *et al.* (2004) have produced reliable and robust systems, but have not overcome the problems of potential damage and slow, impractical convergence times. Parallel evolution with a number of robots (for example Watson *et al.* 1999) reduces the time required, but can still be extremely prohibitive in terms of time and logistics. Simulated robots provide a definite advantage for speed of convergence, but the trade-off is the huge difference between the simulated and real domains (Brooks 1992).

Systems that employ an evolutionary training period (LTL) and some form of lifelong adaptation (STL) have been used to try to address the problem of domain differences, for example, Walker *et al.* (2006) use a GA in the simulated LTL phase and an evolutionary strategy (ES) on the physical robot. They note improved performance when the LTL phase is implemented, and remark that the ES provides continued adaptation to the environment, but they deal with a limited number of behaviour parameters in the GA, and do not state the duration of the LTL phase. Keymeulen *et al.* (1998) run their LTL and STL phases simultaneously, as the physical robot maps its environment at the same time as carrying out its goal-seeking task, thus creating the simulated world. They report the rapid evolution of adaptable and fit controllers, but these results apply only to simple, structured environments where the robot can always detect the coloured target, and the obstacles are few. For example, they observe the development of obstacle avoidance in five minutes, but this applies to an environment with only one obstacle, and the results imply that the real robot was unable to avoid the obstacle prior to this.

The method described here aims to capitalize on the fast convergence speeds that a simulator can achieve, but will also address the domain compatibility issues by transferring the behaviours to an adaptive AIS that runs on a real robot. In theory the method should be entirely practical for real world situations, in terms of delivering a short training-period, safe starting-behaviours, and a fully-dynamic and adaptable system.

The aims of this paper are to investigate whether there are distinct advantages to integrating LTL strategies with STL strategies (for this purpose unseeded systems that use random behaviour sets are also trialed), and to establish the role of the idiotypic network in providing flexibility. The important questions are whether the evolved antibodies can be used effectively in real world environments, or whether there is a need to replace the original antibodies with new ones. In addition, trials using a slightly more complex environmental model are conducted to determine whether this enhances performance. The paper thus aims to investigate the following hypotheses:

$H_1$     Seeded STL systems outperform unseeded STL systems.

$H_2$     Seeded STL systems that employ idiotypic effects outperform seeded systems that rely on RL only.

$H_3$     As long as the LTL-derived behaviours are sufficiently diverse, antibody replacement should not be necessary in the STL phase.

$H_4$     Task performance is enhanced by increasing the number of antigens from eight to nine.

Whitbrook *et al*. (2007) provides statistical evidence that idiotypic AIS systems are more effective than similar non-idiotypic techniques, but this is restricted to a single robotic platform (Pioneer 3), the simulated domain, and only two different environments. The work presented here will hence also extend this research to include a different type of robot (e-puck), more environments, different problems, the real domain, an alternative RL strategy (see sections 4.4 and 5.3), and a variable idiotope (see section 5.1).

## 3   TEST ENVIRONMENTS AND PROBLEMS

The LTL phase requires accelerated simulations in order to produce the initial sets of antibodies as rapidly as possible. For this reason the Webots simulator (Michel 2004) is selected as it is able to run simulations up to 600 times faster than real time, depending on computer power, graphics card, world design and the number and complexity of the robots used. The chosen robot is the e-puck (see Fig 1), since the Webots c++ environment natively supports it. It is a miniature mobile-robot equipped with a ring of eight noisy, nonlinear, infra-red (IR) sensors that can detect the presence of objects up to a distance of about 0.1 m. It also has a small frontal camera and receives the raw RGB values of the images from this. Blob-finding software is created to translate the RGB data into groups of like-coloured pixels (blobs).

The GA runs in a test environment that consists of a virtual e-puck navigating around a building with three rooms (see Figs 2 and 3) by tracking blue markers painted on the walls. These markers are intended to guide the robot through the doors, which close automatically once the robot has passed through. A run ends when the robot has crossed the finish-line in the third room, and its performance is measured according to speed of task completion $^{L}T$, and number of collisions recorded $^{L}C$. Two variations of the test environments are used; World 1 (see Fig 2) has fewer obstacles and no other robots. World 2 (see Fig 3) includes many more obstacles, and there is also a dummy wandering-robot in each room.

The STL is tested in both the virtual and real domains, and the simulated environments are named World 3 and World 4 (see Figs 4 and 5). In these the robot begins south of the central row of pillars and must detect and travel to the blue target-block in the north, avoiding collisions. In addition, a wandering e-puck acts as a dynamic obstacle. Once the robot has arrived at the target, the number of collisions $^{S}C$ and task completion time $^{S}T$ are recorded. The starting positions of the robots and target block are changed automatically after each run.

The real environment consists of a square wooden pen with sides 1.26 m long and 0.165 m high (see Fig 6), and the mission robot must find and travel to a blue ball located inside it, avoiding collisions. Once the ball is found it must come to a complete stop. The obstacles, robots and ball are randomly placed in different starting positions after each run, to create a slightly different environment each time.

A hand-designed controller is also used for comparison with the seeded idiotypic system. This uses a simple random wander for target searching, a backward turning motion to escape collisions, and it steers the robot in the opposite direction to any detected obstacles.

The simulations are run with Webots version 5.1.10 using GNU/Linux 2.6.9 (CentOS distribution) with a Pentium 4 processor (clock speed 3.6 GHz). Fast mode is used for the LTL, and real time for the STL. The graphics card is an NVIDIA GeForce 7600GS, which affords average simulation speeds of approximately 200-times real-time for World 1 and 100-times real-time for World 2. The camera field-of-view is set at 0.3 radians, the pixel width and height at 15 and 3 pixels respectively and the speed unit for the wheels is set to 0.00683 radians/s.

## 4    LONG-TERM LEARNING (GA) SYSTEM ARCHITECTURE

### 4.1    Antigens and Antibodies

The antigens model the environmental information as perceived by the sensors. There are only two basic types of antigen, whether the target is visible (a "target" type) and whether an obstacle is near (an "obstacle" type), the latter taking priority over the former. An obstacle is detected if the IR sensor with the maximum reading $I_{max}$ has value $V_{max}$ equal to 250 or more. If this is the case then the antigen is of type "obstacle", and the antigen is further classified in terms of the obstacle's distance from and its orientation toward the robot. The distance is "near" if $V_{max}$ is between 250 (about 0.03 m) and 2400 (about 0.01 m), and "collision" if $V_{max}$ is 2400 or more. The IR sensors correspond to the quantity of reflected light, so higher readings mean closer obstacles. The orientation is "right" if $I_{max}$ is sensor 0, 1 or 2, "rear" if it is 3 or 4 and "left" if it is 5, 6 or 7 (see Fig 1). If no obstacles are detected then the perceived antigen is of type "target" and there are two varieties, "target seen" and "target unseen", depending on whether appropriate-coloured pixel-clusters have been recognized by the blob-finding software. There are thus eight possible antigens, which are coded 0–7, see Table 1.

Six basic types of behaviour are employed; wandering using either a left or right turn, wandering using both left and right turns, turning forwards, turning on the spot, turning backwards, and tracking the door-markers. Behaviours hence possess an attribute type $U$, and a further six attributes are encoded to enable behaviour diversity. These are speed $S$, frequency of turn $F$, angle of turn $A$, direction of turn $D$, frequency of right turn $R_f$, and angle of right turn $R_a$. The fusion of the basic behaviour-types with a number of attributes that can take many values means that the GA has the potential to select from a huge number of possible robot actions. However, some behaviour types do not use a particular attribute and there are limits to the values that the attributes can take. These limits (see Table 2) are carefully selected in order to strike a balance between reducing the size of the search space, which increases speed of convergence, and maintaining diversity.

Table 1: System antigens

| Antigen Code | Antigen Type | Name |
|---|---|---|
| 0 | Target | Target unseen |
| 1 | Target | Target seen |
| 2 | Obstacle | Obstacle near right |
| 3 | Obstacle | Obstacle near rear |
| 4 | Obstacle | Obstacle near left |
| 5 | Obstacle | Collision right |
| 6 | Obstacle | Collision rear |
| 7 | Obstacle | Collision left |

Table 2: System antibody types

| U | Description | S Speed Units / s | | F % of time | | A % reduction in speed of one wheel | | D Either left or right | | $R_f$ % of time | | $R_a$ % reduction in right wheel-speed | |
|---|---|---|---|---|---|---|---|---|---|---|---|---|---|
| | | \multicolumn{12}{c}{LIMITS} | | | | | | | | |
| 0 | Wander single | 50 | 800 | 10 | 90 | 10 | 110 | L | R | - | - | - | - |
| 1 | Wander both | 50 | 800 | 10 | 90 | 10 | 110 | - | - | 10 | 90 | 10 | 110 |
| 2 | Forward turn | 50 | 800 | - | - | 20 | 200 | L | R | - | - | - | - |
| 3 | Static turn | 50 | 800 | - | - | 100 | 100 | L | R | - | - | - | - |
| 4 | Reverse turn | 500 | 800 | - | - | 20 | 200 | L | R | - | - | - | - |
| 5 | Track markers | 50 | 800 | - | - | 0 | 30 | - | - | - | - | - | - |

### 4.2 GA System Structure

The GA control program uses the two-dimensional array of behaviours $B_{ij}$, $i = 0, \ldots, x-1$, $j = 0, \ldots, y-1$, where $x$ is the number of robots in the population ($x \geq 5$) and $y$ is the number of antigens, i.e. eight. When the program begins *i* is equal to zero, and the array is initialized to null. The infra-red sensors are read every 192 milliseconds, but the camera is only read if no obstacles are found as this increases computational efficiency.

Once an antigen code is determined, a behaviour or antibody is created to deal with it by randomly choosing a behaviour type and its attribute values. For example, the behaviour WANDER_SINGLE (605, 50, 90, LEFT, NULL, NULL) may be constructed. This behaviour consists of travelling forwards with a speed of 605 Speed Units/s, but turning left 50% of the time by reducing the speed of the left wheel by 90%. (Note that wheel speed reductions of more than 100% represent the wheels turning backwards.) The newly created action is executed and the sensor values are read again to determine the next antigen code. If the antigen has been encountered before, then the behaviour assigned previously is used, otherwise a new behaviour is created. The algorithm proceeds in this manner, creating new behaviours for antigens that have not been seen before and reusing the behaviours allotted to those that have. However, the behaviour's cumulative reinforcement-learning score *E*, which is a measure of how well it is thought to have performed, is adjusted after every sensor reading. If *E* falls below the threshold value of -14 then the behaviour is replaced with a new one. Behaviour replacement also occurs when the antigen has not changed in any 60-second period, as this most likely means that the robot has not undergone any translational movement.

A separate supervisor-program is responsible for returning the virtual robot back to its start-point once it has passed the finish-line, for opening and closing the doors as necessary, and for repositioning the wandering dummy-robot, so that it is always in the same room as the mission robot. Another of the supervisor's functions is to assess the time taken to complete the task $^LT$. Each robot is given 1250 seconds to reach the end-point; those that fail receive a 1000-second penalty if they do not pass through any doors. Reduced penalties of 750 or 500 seconds are awarded to failing robots that pass through one door or two doors respectively. When the whole population has completed the course, the relative-fitness $^L\mu$ of each individual is calculated. Since high values in terms of both $^LT$ and $^LC$ should yield a low relative-fitness, the following formula is used:

$$^L\mu_i = \frac{1}{f_i \sum_{k=0}^{x-1}[f_k]^{-1}}, \tag{1}$$

where $^Lf$ is the absolute-fitness given by:

$$^Lf_i = {^LT_i} + \rho\, {^LC_i}. \tag{2}$$

In this phase, $\rho$ is set to 1 to give greater weight to the task time, otherwise robots that constantly turn on the spot, and hence endure no collisions, would receive good relative-fitness values and the GA would take too long to converge.

The five fittest robots from each generation are selected, and their mean task time and mean number of collisions are calculated. The mean absolute-fitness is derived from these using (2) and compared with that of the previous generation to assess rate-of-convergence. The GA terminates when any of the four conditions shown in Table 3 are reached. These are selected in order to achieve fast convergence, but also to maintain a high solution quality. (Note that the convergence criteria are relaxed for World 2, as it is a more cluttered environment requiring a longer task completion time.) If the stopping criteria are met, the attribute values representing the behaviours of the five fittest robots are saved for seeding the AIS system, otherwise the GA proceeds as described in section 4.3.

Note that when adopting the scenario of five separate populations that never interbreed, the five robots that are assessed for convergence are the single fittest from each of the autonomous populations. In this case, convergence is dependent upon the single best $^LT$, $^LC$, and $^Lf$ values, and the final five robots that pass their behaviours to the AIS system are the single fittest from each population after the GA is complete.

Table 3: Stopping criteria. ($g$ = generation number)

| Criteria - World 1 | Criteria - World 2 |
|---|---|
| $g > 0$ AND $^LT_g < 400$ AND $^LC_g < 60$ AND $|^Lf_g - {}^Lf_{g-1}| < 0.1$ | $g > 0$ AND $^LT_g < 600$ AND $^LC_g < 90$ AND $|^Lf_g - {}^Lf_{g-1}| < 0.2$ |
| $g > 30$ | $g > 30$ |
| $^LT_g < 225$ AND $^LC_g < 35$ | $^LT_g < 400$ AND $^LC_g < 45$ |
| $g > 15$ AND $|^Lf_g - {}^Lf_{g-1}| < 0.1$ | $g > 15$ AND $|^Lf_g - {}^Lf_{g-1}| < 0.2$ |

### 4.3 GA Details

Two different parent robots are selected through the roulette-wheel method and each of the $x$ pairs interbreeds to create $x$ child robots, ($x$ is the number of robots in the population). The process is concerned with assigning behaviour attribute-values to each of the child robots for each of the eight antigens in the system. It can take the form of complete antibody replacement, adoption of the attribute values of only one parent or crossover from both parents, and attribute-value mutation.

- Complete antibody replacement occurs according to the prescribed mutation rate $\varepsilon$. Here, a completely new random behaviour is assigned to the child robot for the particular antigen, i.e. both the parent behaviours are ignored.

- Crossover is used when there has been no complete replacement, and the method used depends on whether the parent behaviours are of the same antibody type $U$.
    - If the types are different then the child adopts the complete set of attribute values of one parent only, which is selected at random.
    - If the types are the same, then crossover can occur by taking the averages of the two parent values, by randomly selecting a parent value, or by taking an equal number from each parent according to set patterns. In these cases, the type of crossover is determined randomly with equal probability. The purpose behind this approach is to attempt to replicate nature, where the offspring of the same two parents may differ considerably each time they reproduce.
- Mutation of an attribute value may also take place according to the mutation rate $\varepsilon$, provided that complete replacement has not already occurred. Here, the individual attribute-values (except $D$) of a child robot may be increased or decreased by between 20% and 50%, but must remain within the prescribed limits.

## 4.4  Reinforcement Learning in the Long-Term Learning Phase

Reinforcement Learning (RL) is used to accelerate GA convergence, and works by comparing current and previous antibody codes to determine behaviour effectiveness. Ten points are awarded for every positive change in the environment, and ten are deducted for each negative change. Table 4 shows the possible antigen code combinations and column 3 shows the points added or deducted in the LTL-phase. For example, 20 points are awarded if the antigen code changes from an "obstacle" type to "target seen", because the robot has moved away from an obstacle as well as gaining or keeping sight of the target. In the case where the antigen code remains at 1 (the target is kept in sight), the score awarded depends upon how the orientation of the target has moved with respect to the robot. In addition, when an obstacle is detected both in the current and previous iteration, then the score awarded depends upon several factors, including changes in the position of $I_{max}$ and in the reading $V_{max}$, the current and previous distance-type ("collision" or "near") and the tallies of consecutive "nears" and "collisions". Further details on the LTL architecture are provided in Whitbrook *et al.* (2008a).

Table 4: Reinforcement scores in the LTL and STL phases

| Antigen Code | | Score (LTL) | Score (STL) | Reinforcement status (score) |
|---|---|---|---|---|
| Previous | Current | | | |
| 0 | 0 | 0 | 0.05 | Neutral |
| 1 | 0 | -10 | -0.10 | Penalize - Lost sight of target |
| 2-7 | 0 | 10 | 0.10 | Reward - Avoided obstacle |
| 0 | 1 | 10 | 0.10 | Reward - Found target |
| 1 | 1 | 0 - 5 | 0.00- 0.05 | Reward – Kept sight of target (Score depends on orientation of target with respect to robot) |
| 2-7 | 1 | 20 | 0.20 | Reward - Avoided obstacle and gained or kept sight of target |
| 0 | 2-7 | 0 | -0.05 | Neutral |
| 1 | 2-7 | 0 | -0.05 | Neutral |
| 2-7 | 2-7 | -4 - 5 | -0.40 -0.50 | Reward or Penalize (Score depends on several factors) |

# 5 SHORT-TERM LEARNING (AIS) SYSTEM ARCHITECTURE

## 5.1 Creating the Paratope and Idiotope Matrices

The GA selects the five fittest robots from the final generation, so five distinct sets of antibodies are used, each set consisting of eight behaviours, i.e. one antibody for each antigen. The 40 antibodies in the system can hence be represented as $A_{ij}$, $i = 0, \ldots, v\text{-}1$, $j = 0, \ldots, y\text{-}1$, where $v$ is the number of sets and $y$ is the number of antigens. The evolved antibody types and their associated attribute values, task completion times $^{L}T_i$ and numbers of collisions $^{L}C_i$ are taken directly from the file created in the LTL phase. The STL phase calculates the relative fitness of each antibody set $^{S}\mu_i$ from:

$$^{S}\mu_i = \frac{1}{f_i \sum_{k=0}^{v-1} [f_k]^{-1}}, \qquad (3)$$

using (2) with $\rho$ set to 8 to give the collisions approximate equal weight compared to the task time. (This is permissible here because it is assumed that evolution will not have selected robots that constantly turn on the spot.) Once the relative fitness values are calculated, a matrix of RL scores $P_{ij}$ can be derived by multiplying the antibody's final RL score $E_{ij}$ by the relative fitness $^{S}\mu_i$ of its set, and scaling approximately to between 0.00 and 1.00 using:

$$P_{ij} = \frac{E_{ij} {}^{S}\mu_i}{\varphi}. \qquad (4)$$

Taking $\varphi$ as 20 achieves the required scaling in (4) since the maximum value $E_{ij}{}^S\mu_i$ can take is approximately 20. The matrix $P$ is analogous to an antibody paratope as the scores represent a comparative estimate of how well each antibody matches its antigen.

For the unseeded systems the five antibody sets are generated at the start of the STL phase, by randomly choosing behaviour types and their attribute values. The initial elements of $P$ are also randomly generated, but always lie between 0.25 and 0.75 to try to limit any initial biasing of the selection.

For both seeded and unseeded systems, a matrix $I$ (analogous to a matrix of idiotope values) is created by comparing the individual paratope matrix elements $P_{ij}$ with the mean element value for each of the antigens $\sigma_j$. This is given by:

$$\sigma_j = \frac{\sum_{i=0}^{v-1} P_{ij}}{v}. \tag{5}$$

If $P_{ij}$ ($i = 0, \ldots, v$-1) is less than $\sigma_j$, then an idiotope value $I_{ij}$ of 1.0 is assigned, otherwise a value of zero is given. However, only one antibody in each set may have a non-zero idiotope. If more than one has a non-zero value, then one is selected at random and the others are set back to zero. This avoids over-stimulation or over-suppression of antibodies.

The paratope matrix is adjusted after every iteration; first, because the active antibody's paratope value either increases or decreases, depending on the RL score awarded, and second, because the paratope values are re-calculated, so that each $\sigma_j$ is returned to its initial mean value. The adjustment is given by:

$$P_{ij_{t+1}} = P_{ij_t} \frac{\sigma_{j_0}}{\sigma_{j_t}}, \tag{6}$$

where $\sigma_{j0}$ represents the initial mean and $\sigma_{jt}$ represents the temporary mean obtained after scoring of the active antibody. The adjustment helps to eliminate the problems that occur when useful antibodies acquire zero $P_{ij}$ values. The idiotope is recalculated, based on the latest $P_{ij}$ values, after every 120 sensor readings.

**5.2 Antibody Selection Process**

At the start of the STL phase each antibody has 1000 clones in the system, but the numbers fluctuate according to a variation of Farmer's equation:

$$N_{im_{(t+1)}} = bS_{im_{(t)}} + N_{im_{(t)}}(1 - k_3), \tag{7}$$

where $N_{im}$ represents the number of clones of each antibody matching the invading antigen $m$, $S_{im}$ is the current strength-of-match of each of these antibodies to $m$, $b$ is a scaling constant and $k_3$ is the death rate constant. The concentration $C_{ij}$ of every antibody in the system consequently changes according to:

$$C_{ij} = \frac{\Phi N_{ij}}{\sum_{k=0}^{x-1}\sum_{l=0}^{y-1} N_{kl}}, \tag{8}$$

where $\Phi$ is another scaling factor that can be used to control the levels of inter-antibody stimulation and suppression (25 is used here).

The antibody selection process comprises three stages for idiotypic selection, but only one stage if idiotypic selection is not used. First, the sensors are read to determine the index of the presenting antigen $m$, and an appropriate antibody is selected from those available for that antigen. More specifically, the system chooses from antibodies $A_{im}$, $i = 0, \ldots, 4$, by examining the paratope values $P_{im}$. The antibody $\alpha$ with the highest of these paratope values is chosen as the first-stage winner. If the index of the winning antibody set is denoted as $n$, then $\alpha = A_{nm}$. If idiotypic effects are not considered $\alpha$ carries out its action, and is assessed by RL, see section 5.3.

If an idiotypic system is used, then the stimulatory and suppressive effects of $\alpha$ on all the antibodies in the repertoire must be considered. This involves comparing the idiotope of $\alpha$ with the paratopes of the other antibodies to determine how much each is stimulated, and comparing the paratope of $\alpha$ with the idiotopes of the others to calculate how much each should be suppressed. Here, idiotypic selection is governed by equations (9)-(12), which are based on those in Whitbrook *et al.* (2007). Equation (9) concerns the increase in strength-of-match value $\varepsilon_{im}$ when stimulation occurs,

$$\varepsilon_{im} = k_1 \sum_{j=0}^{y-1} (1 - P_{ij}) I_{nj} C_{ij} C_{nj}, \tag{9}$$

where $k_1$ is a constant that determines the magnitude of any stimulatory effects. The formula for the reduction in strength-of-match value $\delta_{im}$ when suppression occurs is given by:

$$\delta_{im} = k_2 \sum_{j=0}^{y-1} P_{nj} I_{ij} C_{ij} C_{nj}, \tag{10}$$

where $k_2$ governs the suppression magnitude. Hence, the strength-of-match after the second selection-stage $(S_{im})_2$ is given by:

$$(S_{im})_2 = (S_{im})_1 + \varepsilon_{im} - \delta_{im}, \qquad (11)$$

where the initial strength-of-match $(S_{im})_1$ for each antibody is taken as the current $P_{im}$ value. After the $(S_{im})_2$ values are calculated, the numbers of clones $N_{im}$ are adjusted using (7) and all concentrations $C_{ij}$ are re-evaluated using (8). The third stage calculates the activation $\lambda$ of each antibody in the sub-set $A_{im}$ from:

$$\lambda_{im} = C_{im}(S_{im})_2. \qquad (12)$$

The third-stage winning antibody $\beta$ is that with the highest $\lambda$ value in the sub-set. If $p$ is the index of $\beta$'s antibody set, then $\beta = A_{pm}$. When idiotypic selection is used, $\beta$ carries out its action and it is $\beta$ that is scored using RL rather than $\alpha$, although $\alpha$ and $\beta$ are the same when $n = p$.

**5.3 Reinforcement Learning within the Short-Term Learning Phase**

In the STL phase the RL scores are scaled to one hundredth of the values used for the LTL phase (see column 4 of Table 4), since the RL is intimately linked with the idiotypic selection process, and larger values would lead to over-stimulation and over-suppression. In addition, a reward is given when no obstacles are encountered, and penalties are issued when they are. This is in contrast to the LTL case, where no reward or penalty is issued, and is necessary to increase the flux of the system. In the

LTL, neutral scores are permissible as there is ample time to develop good strategies, but in the STL, the idiotypic system needs to remain in a state of flux if suppression and stimulation are to occur at all.

The maximum cumulative-RL-score (or $P_{ij}$ value) allowed is 1.00, and the minimum $P_{ij}$ value is 0.00. The $P_{ij}$ values are also adjusted when the antigen code has remained at 0 for more than 250 iterations, as this means that the robot is spending too much time wandering and has not found anything. It is important to recognize this behaviour as negative, as otherwise robots may be circling around on the spot, never achieving anything, but receiving constant rewards. The non-idiotypic case reduces the cumulative-RL-score by 1.0, and the idiotypic case reduces it by 0.5, as pre-trials have shown that non-idiotypic robots require a more drastic change to break out of repeated behaviour cycles. The same $P_{ij}$ adjustments are also made if there have been more than 15 consecutive obstacle encounters, as this may indicate that a robot is trapped. Following RL, the paratope values are scaled using (6).

In the case of the unseeded trials, replacement occurs for all antibodies with a cumulative- RL-score less than 0.1. The successor is created by randomly choosing a behaviour type and its attribute values. Replacement does not occur in the seeded systems, since $H_3$ is directly concerned with establishing whether this is necessary. Further details on the STL architecture are provided in Whitbrook *et al*. (2008b).

## 6    EXPERIMENTAL PROCEDURES AND RESULTS

### 6.1    Long-Term Learning General Procedures

The GA is run in Worlds 1 and 2 using single populations of 25, 40, and 50 robots, and using five autonomous populations of five, eight, and ten. A mutation rate $\varepsilon$ of 5% is used throughout, as previous trials have shown that this provides a good compromise between fast convergence, high diversity and good solution-quality. Solution quality $^{L}q$ is taken as half of the absolute-fitness value (2) with $\rho = 8$, to give approximate equal weighting to the collisions. For each scenario, ten repeats are performed and the means of the convergence time $\tau$, solution quality $^{L}q$, and diversity in type $Z_U$ and speed $Z_S$ (see Section 6.2) are recorded. Two–tailed standard $t$-tests are conducted on the result sets, and differences are accepted as significant at the 99% level only.

### 6.2    Measuring Antibody Diversity

Antibody diversity is measured using the type $U$ and the speed $S$ attributes, since these are the only action-controlling attributes common to all behaviours. The final antibodies are grouped by antigen number and the groups are assessed by comparison of each of the five members with the others, i.e. ten pair-wise comparisons are made in each group. A point is awarded for each comparison if the attribute values are different; if they are the same no points are awarded. For example, the set of behaviour types [1 3 4 4 1] has two pair-wise comparisons with the same value, so eight points are given. Table 5 summarizes possible attribute-value combinations and the result of conducting the pair-wise comparisons on them.

The $y$ individual diversity-scores for each of $U$ and $S$ are summed and divided by $\sigma y$ to yield a diversity score for each attribute. Here $\sigma$ is the expected diversity-score for a large number of randomly-selected sets of five antibodies. This is approximately 8.333 for $U$ (see Table 5) and 10.000 for $S$. It is lower for $U$ since there are only six behaviours to select from, whereas the speed is selected from 751 possible values, so there is a much higher probability of producing unique values in a random selection of five. The adjustment effectively means that a random selection yields a diversity of 1 for both $S$ and $U$. The diversity calculation is given by:

$$Z = \frac{\sum_{i=1}^{y} z_i}{\sigma y}, \qquad (13)$$

where $Z$ represents the overall diversity-score and $z$ represents the individual score awarded to each antigen.

Table 5: Diversity scores

| Attribute-value status | Points | Expected: Frequency for $U$ | Expected: Score for $U$ |
|---|---|---|---|
| All five different | 10 | 9.26 | 0.926 |
| One repeat of two | 9 | 46.30 | 4.167 |
| Two repeats of two | 8 | 23.15 | 1.852 |
| One repeat of three | 7 | 15.43 | 1.080 |
| Two repeats, one of two, one of three | 6 | 3.86 | 0.231 |
| One repeat of four | 4 | 1.93 | 0.077 |
| All five the same | 0 | 0.08 | 0.000 |
| **Total** | | **100.00** | **8.333** |

## 6.3 Long-Term Learning Phase Results

Table 6 presents mean $\tau$, $^Lq$, $Z_U$, and $Z_S$ values, and Table 7 summarises the significant difference levels when comparing single and multiple populations of robots. The schemes that are compared use the same number of robots, for example a single population of 25 is compared with five populations of five.

Table 6: Mean values

| Pop. size | World 1 | | | | World 2 | | | |
|---|---|---|---|---|---|---|---|---|
| | $\tau$ (s) | $^Lq$ | $Z_U$ (%) | $Z_S$ (%) | $\tau$ (s) | $^Lq$ | $Z_U$ (%) | $Z_S$ (%) |
| 25 | 417 | 220 | 40 | 86 | 972 | 314 | 37 | 85 |
| 40 | 530 | 216 | 53 | 95 | 1292 | 266 | 51 | 89 |
| 50 | 811 | 191 | 49 | 90 | 1414 | 250 | 56 | 94 |
| 5 x 5 | 508 | 155 | 55 | 100 | 1211 | 258 | 58 | 100 |
| 5 x 8 | 590 | 146 | 54 | 100 | 1325 | 225 | 55 | 100 |
| 5 x 10 | 628 | 144 | 58 | 100 | 1498 | 208 | 57 | 100 |

Table 7: Significant differences

| Comparison | | World 1 | | | | World 2 | | | |
|---|---|---|---|---|---|---|---|---|---|
| | | $\tau$ (s) | $^Lq$ | $Z_U$ (%) | $Z_S$ (%) | $\tau$ (s) | $^Lq$ | $Z_U$ (%) | $Z_S$ (%) |
| 25 | 5 x 5 | 77.40 | **99.94** | **99.90** | **99.99** | 88.47 | 84.51 | **99.96** | **99.63** |
| 40 | 5 x 8 | 72.58 | **99.97** | 43.07 | **99.97** | 20.91 | 94.09 | 61.19 | **99.28** |
| 50 | 5 x 10 | 97.13 | **99.80** | 98.36 | **99.96** | 40.78 | 97.31 | 18.34 | **99.87** |

The tables show that, for both worlds, there are no significant differences between convergence times when comparing the single and multiple populations. In addition, speed diversity is significantly better for the multiple populations in all cases. Multiple populations always demonstrate a speed diversity of 100%, indicating that the final-selected genes are completely unrelated to each other, as expected. In contrast, single-population speed-diversity never reaches 100% as there are always repeated genes in the final-selected robots. Evidence from previous experiments with single populations of five, ten and 20 suggests that the level of gene duplication decreases as the single population size increases. This explains the lower $Z_U$ and $Z_S$ values for a population of 25 robots.

Type diversity is consistently higher for the multiple populations, but only significantly higher when comparing a single population of 25 with five populations of five robots. For the multiple populations, mean type-diversity ratings never reach 100%, even though there are no repeated genes. The reduced type-diversity must occur because there are only six types to choose from, and these are not randomly selected but chosen in a more intelligent way. However, speed diversity can remain at 100% because there are many different speeds to choose from and convergence is rapid. It is likely that both intelligent selection and repeated genes decrease the type-diversity scores for the single populations, but in the multiple populations, the phenomenon is caused by intelligent selection only.

In World 1, solution quality is consistently significantly better for the multiple populations, but this is not the case in World 2. This may indicate that using multiple populations helps to improve solution quality for simpler problems, but the phenomenon diminishes as the problem becomes more difficult.

The best option in terms of population model appears to be five autonomous populations, since this elicits significantly-higher antibody diversity. In addition, one can run the GA without significantly increasing the convergence time or reducing solution quality, and the fast convergence times (ten minutes in World 1 and 25 minutes in World 2) satisfy the requirement for a practical training-period.

## 6.4   Short-Term Learning General Procedures

Pre-trials have shown that the antibody sets taken from the above LTL experiments produce a higher number of collisions compared with a hand-designed controller, when used to seed the AIS in Worlds 3 and 4. Since the hand-designed controller deals with much slower speeds, the GA is run again in World 1 with five autonomous populations of ten robots and with the static-turn antibody's upper speed-limit reduced to 100 speed units/s, all other speed limits reduced to 400 speed units/s, and the reverse antibody's lower speed-limit reduced to 300 speed units/s. The stopping criteria is also simplified to $g > 0$ AND $^LT_g < 500$ AND $^LC_g < 25$ OR $g > 30$ to allow for the general increase in task time.

Thirty STL trials are performed in each of the two simulated worlds, World 3 and World 4, and 20 are completed in the real world. This is done for each of the following systems; seeded with idiotypic effects, seeded with RL only, unseeded with idiotypic effects, unseeded with RL only, and the hand-designed controller. In the unseeded simulated-worlds two separate sets of experiments are conducted with two different initially-random behaviour sets $R_1$ and $R_2$. The real-world unseeded experiments use only $R_1$ since they have to run in real time and are hence much more time consuming to carry out.

In the idiotypic systems $b$ is set to 100, $k_3$ is set to zero, and $k_1$ and $k_2$ are set at 0.85 and 1.10 respectively. These values are chosen in order to yield a mean idiotypic difference rate of approximately 20%, as suggested in Whitbrook *et al.* (2007). Note that an idiotypic difference occurs when the antibodies $\alpha$ and $\beta$ are different.

A run finishes when the robot has detected three consecutive instances of more than 40 blue pixels in the ball image, so that it is "aware" of having found its target. For all experiments, the time taken $^S T$ and the number of collisions $^S C$ are capped at 4000 s and 100 respectively. Any runs that exceed either of these limits are counted as failures. The solution quality, $^S q$ is calculated in the same way as for the LTL, i.e:

$$^S q = \frac{^S T + \rho\, ^S C}{2} \qquad (14)$$

where $\rho = 8$ as before. Standard two-tailed *t*-tests are applied to compare the various systems, and differences are accepted as significant at the 99% level only.

## 6.5  Short-Term Learning Phase Results

Table 8 shows the mean $^S C$, $^S T$, and $^S q$ values for each of the systems in each of the worlds, and Table 9 presents the significant difference levels when the systems are compared. Table 10 highlights the failure rates, indicating the percentage of failures due to an excessive number of collisions, running out of time, and overall.

Table 8. Mean $^S C$, $^S T$, and $^S q$. (S = seeded, U = unseeded, IE = idiotypic effects, RL = reinforcement learning, HDC = hand-designed controller)

| System | Set | Simulated World 3 | | | Simulated World 4 | | | Real World | | |
|---|---|---|---|---|---|---|---|---|---|---|
| | | $^S C$ | $^S T$ | $^S q$ | $^S C$ | $^S T$ | $^S q$ | $^S C$ | $^S T$ | $^S q$ |
| SIE | - | 1 | 562 | 284 | 2 | 659 | 336 | 5 | 283 | 161 |
| SRL | - | 8 | 1298 | 679 | 4 | 1113 | 573 | 23 | 904 | 544 |
| UIE | $R_1$ | 26 | 1513 | 862 | 26 | 1530 | 868 | 96 | 1384 | 1074 |
| URL | $R_1$ | 45 | 2150 | 1253 | 35 | 1732 | 1006 | 100 | 1678 | 1239 |
| UIE | $R_2$ | 20 | 1720 | 941 | 48 | 1578 | 981 | - | - | - |
| URL | $R_2$ | 35 | 2214 | 1246 | 54 | 2137 | 1285 | - | - | - |
| HDC | - | 2 | 1362 | 688 | 2 | 1256 | 636 | 44 | 1439 | 897 |

In all of the worlds, both simulated and real, the seeded idiotypic system proves better in terms of fewer collisions, a faster completion time, and a higher solution quality.

When compared with the unseeded systems it is significantly better in all cases, i.e. for all of the metrics, in all the worlds, and irrespective of whether the unseeded systems use idiotypic effects, or which random behaviour set is used.

Table 9. Significance Levels (S = seeded, U = unseeded, IE = idiotypic effects, RL = reinforcement learning, HDC = hand-designed controller)

| Systems | | Set | Simulated World 3 | | | Simulated World 4 | | | Real World | | |
|---|---|---|---|---|---|---|---|---|---|---|---|
| | | | $^SC$ | $^ST$ | $^Sq$ | $^SC$ | $^ST$ | $^Sq$ | $^SC$ | $^ST$ | $^Sq$ |
| SIE | SRL | - | 100 | 100 | 100 | 98 | 96 | 97 | 99 | 99 | 100 |
| SIE | HDC | - | 85 | 100 | 100 | 33 | 97 | 97 | 100 | 100 | 100 |
| SIE | UIE | $R_1$ | 100 | 100 | 100 | 100 | 100 | 100 | 100 | 100 | 100 |
| SIE | URL | $R_1$ | 100 | 100 | 100 | 100 | 100 | 100 | 100 | 100 | 100 |
| SIE | UIE | $R_2$ | 99 | 100 | 100 | 100 | 100 | 100 | - | - | - |
| SIE | URL | $R_2$ | 100 | 100 | 100 | 100 | 100 | 100 | - | - | - |
| SRL | UIE | $R_1$ | 98 | 49 | 72 | 99 | 83 | 92 | 100 | 85 | 99 |
| SRL | URL | $R_1$ | 100 | 99 | 100 | 100 | 94 | 98 | 100 | 96 | 100 |
| SRL | UIE | $R_2$ | 91 | 82 | 89 | 100 | 86 | 98 | - | - | - |
| SRL | URL | $R_2$ | 100 | 99 | 100 | 100 | 100 | 100 | - | - | - |
| UIE | URL | $R_1$ | 87 | 90 | 93 | 59 | 44 | 52 | 68 | 53 | 57 |
| UIE | URL | $R_2$ | 82 | 81 | 87 | 40 | 86 | 84 | - | - | - |

Table 10. Percentage Failure Rates (S = seeded, U = unseeded, IE = idiotypic effects, RL = reinforcement learning, HDC = hand-designed controller)

| System | Set | Simulated World 3 (%) | | | Simulated World 4 (%) | | | Real World (%) | | | Mean (%) | | |
|---|---|---|---|---|---|---|---|---|---|---|---|---|---|
| | | $^SC$ | $^ST$ | Tot | $^SC$ | $^ST$ | Tot | $^SC$ | $^ST$ | Tot | $^SC$ | $^ST$ | Tot |
| SIE | - | 0 | 0 | 0 | 0 | 0 | 0 | 0 | 0 | 0 | 0 | 0 | 0 |
| SRL | - | 0 | 3 | 3 | 0 | 7 | 7 | 10 | 5 | 10 | 3 | 5 | 7 |
| UIE | $R_1$ | 23 | 17 | 30 | 20 | 13 | 23 | 95 | 10 | 95 | 46 | 13 | 49 |
| URL | $R_1$ | 43 | 30 | 57 | 33 | 23 | 47 | 100 | 20 | 100 | 59 | 24 | 68 |
| UIE | $R_2$ | 17 | 20 | 37 | 43 | 17 | 43 | - | - | - | 30 | 18 | 40 |
| URL | $R_2$ | 30 | 30 | 47 | 50 | 27 | 53 | - | - | - | 40 | 28 | 50 |
| HDC | - | 0 | 20 | 20 | 0 | 17 | 17 | 10 | 25 | 35 | 3 | 21 | 24 |

The seeded idiotypic system also surpasses the hand-designed controller in all cases (except for a tie in $^SC$ in World 4), and more than half of these differences are significant overall. Moreover, in the real world all of the differences are significant. It appears that the hand-designed controller performs very well in the simulator in terms of $^SC$, but poorly for $^ST$, whereas in the real world it performs badly for both of these metrics. Although it has built-in initial knowledge, it probably proves inferior in the real world because of its inability to change the way it responds to an antigen. The seeded idiotypic system works well in the real world and in the simulator for both $^SC$ and $^ST$. In fact, in the real world it proves significantly better than all of the other systems trialled, for all metrics.

When the non-idiotypic seeded system is compared with the unseeded systems, although its performance is better in all cases, it is not always significantly better. Most of the significant differences arise when comparing seeded and unseeded systems that do not use idiotypic effects. When the unseeded system employs idiotypic effects and the seeded system does not, there is a marked drop in the percentage of significant differences.

When the seeded idiotypic system is compared with the seeded non-idiotypic system, the idiotypic system performs better in all cases, and significantly better in most. However, when the unseeded systems are compared in this way, although the idiotypic system consistently performs better, none of the differences are significant.

The seeded idiotypic system is the only scheme that displays an overall failure rate of 0%. Failure rates are reasonably low (7% overall) for the non-idiotypic seeded system, but reach unacceptable proportions for the hand-designed controller (24% overall) and the idiotypic unseeded system (49% and 40% overall). The non-idiotypic unseeded system is clearly the worst option with overall fail rates of 68% and 50%. Moreover, the actual number of collisions for failing robots is of the order of thousands for unseeded real-world systems, which renders the method entirely unsuitable.

A general observation is that both the hand-designed controller and the non-idiotypic seeded system exhibit repeated behaviour patterns, particularly when obstacles are both to the left and to the right of the robot. Under these circumstances the robot often moves away from one obstacle, only to encounter the other, and the sequence continues, sometimes indefinitely. The phenomenon is also observed with the seeded idiotypic system, but to a much lesser extent, and the robot is always able to free itself quite promptly.

### 6.6     Representation of the Antigen Space

The seeded idiotypic robot still sometimes exhibits a repeated behaviour pattern when there are objects both to its left and right. Although this is observed rarely, and the robot is always able to free itself quite quickly, it may be that the antigen coverage is not represented adequately. This raises the question as to whether there are potential benefits to introducing an additional environmental scenario "obstacle left and right". Its inclusion might also improve the performance of the non-idiotypic and hand-designed systems, reducing or even eliminating any idiotypic advantage. In order to investigate these matters, a single new antigen is created, and the antigens are recoded as shown in Table 11. The new antigen (coded 5) presents itself when IR sensors 5 and 2 (those directly to the left and right of the robot respectively) both exhibit readings between 140 and 2400. However, if $V_{max}$ is above 2400, antigen 6, 7, or 8 (i.e. one of the collision antigens) is invoked, as before. There is no "collision left and right" antigen as a series of simulated and real world trials suggests that it simply does not appear. In addition, the mixed antigens "object near left and collision right", and "collision left and object near right" are not included as they occur very rarely and would produce too many new antigens, probably increasing the execution time of the LTL phase too much.

Pre-trials also show that, when the previous antigen is the newly introduced one (i.e. antigen 5) and any object is detected, it is necessary to use only positive RL scores, and to boost them. This is because the new antigen does not occur as often as the others, and so any negative RL scores effectively put the assigned behaviours out of business very quickly. The required changes are accomplished by ignoring the actual $V_{max}$ values and the object distances, and by making all scores awarded for changes in orientation positive and tripling them, see Table 12.

Table 11: Recoded system antigens

| New Antigen Code | Antigen Type | Name |
|---|---|---|
| 0 | Target | Target unseen |
| 1 | Target | Target seen |
| 2 | Obstacle | Obstacle near right |
| 3 | Obstacle | Obstacle near rear |
| 4 | Obstacle | Obstacle near left |
| 5 | Obstacle | Obstacle near left and right |
| 6 | Obstacle | Collision right |
| 7 | Obstacle | Collision rear |
| 8 | Obstacle | Collision left |

Table 12. Reinforcement scores with the additional antigen in the STL phase

| New Antigen Code | | Score | Reinforcement status |
|---|---|---|---|
| Previous | Current | | |
| 0 | 0 | 0.05 | Reward – No obstacles encountered |
| 1 | 0 | -0.10 | Penalize - Lost sight of target |
| 2-8 | 0 | 0.10 | Reward - Avoided obstacle |
| 0 | 1 | 0.10 | Reward - Found target |
| 1 | 1 | 0.00 to 0.05 | Reward – Kept sight of target (Score depends on orientation of target with respect to robot) |
| 2-8 | 1 | 0.20 | Reward - Avoided obstacle and gained or kept sight of target |
| 0 | 2-8 | -0.05 | Penalize – Encountered obstacle |
| 1 | 2-8 | -0.05 | Penalize – Encountered obstacle |
| 2-4, 6-8 | 2-8 | -0.40 to 0.50 | Reward or Penalize (Score depends on several factors) |
| 5 | 2-4, 6-8 | 0.00 to 0.54 | Reward or Penalize (Score depends on several factors) |
| 5 | 5 | 0.00 to 0.54 | Reward or Penalize (Score depends on several factors) |

Following these adjustments, the GA is re-run with five autonomous populations of ten, as before, in order to obtain the new sets of starting antibodies. The STL experimental procedures used for the eight-antigen structure are repeated for the nine-antigen structure, except that only $R_1$ is used for the unseeded systems. When compared with the results from the eight-antigen structure, the new results show no significant difference for the seeded and unseeded idiotypic systems, but there are some significant differences for the non-idiotypic schemes. With nine antigens the seeded non-idiotypic system performs significantly better for all of the metrics in World 3, and the unseeded non-idiotypic network performs significantly better for collisions in World 3, but significantly worse for time and solution quality in the real world.

These results translate to the following changes when comparing the nine-antigen systems with each other:

- The seeded idiotypic system is still always significantly better than the unseeded systems in terms of time and solution quality, but in the simulator the collisions show less significance now.
- The seeded idiotypic system still consistently out performs the seeded non-idiotypic system, but the only significant differences are in World 4 for time and solution quality.
- The unseeded idiotypic system is now mostly significantly better than the unseeded non-idiotypic system.

## 6.7 Discussion

The observations detailed in section 6.5 provide very strong statistical evidence in support of $H_1$, i.e. they defend the notion that seeded schemes outperform unseeded ones. The results also uphold $H_2$, since robot performance appears to be further enhanced by incorporating an idiotypic network into the STL architecture. In the seeded idiotypic system, the evolved antibody set provides immediate knowledge of how to begin the task, and the idiotypic AIS permits it to change and adapt its behaviour as the need arises. Without idiotypic effects, the seeded system has the same initial knowledge, but relies only on RL for adaptation, so it is less flexible. However, when the unseeded systems are compared in this way, no significant difference is apparent. This is because the unseeded systems have no initial knowledge, and must acquire their abilities during the STL phase. This is a very slow process, even when idiotypic selection is used, because the search space is probably much too large given the time frame for completing the task. Moreover, the mechanism by which antibodies are replaced is not well developed; the robot is forced to select a random behaviour when it rejects an antibody, and could hence still be using random antibodies during the latter stages of task completion.

Further evidence in favour of coupling LTL seeding with STL idiotypic mechanisms lies in the fact that the seeded idiotypic system is the only scheme that consistently displays a 0% failure rate. This upholds $H_3$, i.e. it suggests that antibody replacement is not necessary when adequate seeding and a sufficiently adaptive strategy are in place.

The effect of the extra antigen is to reduce the number of collisions for the unseeded systems, but only in simulation. It also brings about reduced numbers of collisions, and an all-round better performance for the seeded non-idiotypic systems, especially for the simpler simulated world, but the improvement is not consistent throughout all the environments, and the system still exhibits an overall fail rate of 3% compared with 0% again for the seeded idiotypic system. As the collision reduction translates poorly to the real world, and the improvement in non-idiotypic systems is inconsistent, there is not much support for $H_4$, and so use of the extra antigen is not recommended. In addition, its use increases the convergence time of the GA, and based on performance, one would always opt for the seeded idiotypic system, which shows no significant change when the new antigen is introduced.

## 7  CONCLUSIONS

This paper has described merging LTL (an accelerated GA), with STL (an idiotypic AIS), in order to seed the AIS with sets of very diverse behaviours that can work together to solve a mobile-robot target-finding problem. It has described the unique antibody encoding and the GA method used for evolving the initial set of antibodies, and has shown that significantly higher antibody diversity can be obtained when a number of autonomous populations are used, rather than a single one. Furthermore, for five autonomous populations, one can run the GA without significantly increasing convergence time or reducing solution quality, and the diversity ratings do not appear to be affected by the difficulty of the problem. The LTL system has proved itself capable of delivering the starting antibodies within a realistic time frame, i.e. within about ten minutes in a static world, and within about 25 minutes in a dynamic world.

The STL phase architecture has also been described and a number of experiments performed that show that seeded systems consistently perform significantly better than unseeded systems in both the real world and different simulated worlds. Strong statistical evidence that the idiotypic selection process contributes towards this improvement has also been demonstrated, and the experiments further imply that antibody replacement is not necessary within the STL-phase as long as adequate seeding is in place. In addition, trials have been conducted with an extra antigen, but these have shown no significant benefit, suggesting that eight antigens may already be optimal in terms of balancing LTL convergence-time and STL performance.

The fusion of the two learning timescales has hence provided an adaptable and robust system for carrying out navigation activities in structured real-world environments. This shows that, given the right conditions, behaviours derived in GA simulations can transfer extremely well to the real world, even when the nature and layout of the environments are quite different.

## ACKNOWLEDGEMENTS

This work was funded by the UK Government's Engineering and Physical Sciences Research Council (EPSCRC).

FIGURE CAPTIONS

Fig 1 An e-puck robot showing IR sensor positions and frontal camera.

Fig 2 Simulated World 1

Fig 3 Simulated World 2

Fig 4 Simulated World 3

Fig 5 Simulated World 4

Fig 6 The real world environment